# A Simple Generative Network

Daniel N. Nissani (Nissensohn)
dnissani@post.bgu.ac.il

**Abstract.**

Generative neural networks are able to mimic intricate probability distributions such as those representing handwritten text, natural images, etc. Since their inception several models were proposed. The most successful of these were based on adversarial (GAN), auto-encoding (VAE) and maximum mean discrepancy (MMD) relatively complex architectures and schemes. Surprisingly, a very simple architecture (a single feed-forward neural network) in conjunction with an obvious optimization goal (Kullback–Leibler divergence) was apparently overlooked. This paper demonstrates that such a model (denoted SGN for its simplicity) is able to generate samples visually and quantitatively competitive as compared with the fore-mentioned state of the art methods.

## 1 Introduction

This work is about generative neural networks. When fed by input samples from some (typically simple) distribution, these networks are able to generate output samples which mimic data obeying arbitrary probability distributions, such as those representing natural images, handwritten text, etc. Since the publication of Variational Autoencoders (VAE, Kingma and Welling, 2014) several other similarly capable models have been proposed. These include Generative Adversarial Networks (GAN, Goodfellow et al., 2014), Maximum Mean Discrepancy networks (MMD, Li et al. 2015; Dziugaite et al., 2015) and Normalizing Flows networks (Rezende and Mohamed, 2015). For a comparative exhaustive survey (and even more exhaustive references list) the reader may refer to (Mohamed and Lakshminarayanan, 2017; Bond-Taylor et al., 2021).

With such a variety of schemes it is curious and surprising that a most simple architecture and direct approach seems to have been overlooked. It is the modest purpose of this paper to fill this gap.

Amongst the approaches cited above the one conceptually closest to our work is the MMD network (Li et al. 2015). The architecture of this model is very simple and consists of a single feed-forward multi-layer neural net. This stands in contrast to the popular fore-mentioned GANs which are based on two competing networks, a generator and a discriminator, and to the VAEs which are based on an auto-encoding fer from delicate convergence issues; and VAEs on the other hand have apparently not been able so far to generate high fidelity samples. The MMD (a.k.a. GMMN; Li et al. 2015) model is based on the well known principle (e.g. Mnatsakanov, 2008) which states that, under relatively mild conditions, two distributions which are equal in all their moments are themselves equal. This principle led the authors to adopt a (Gaussian) kernel functions based optimization goal. This choice in turn yielded poor generated samples, both in terms of visual quality and in terms of Mean Log Likelihood (MLL), a popular quantitative measure. As result of this the authors of this work shifted their focus to a modified variant of MMD networks (denoted GMMN + AE), based on a more complex auto-encoding architecture.

We adopt in our present work the single network simple architecture and will demonstrate in the sequel that it is possible to generate high quality samples by a proper, different choice of optimization goal, namely the Kullback–Leibler divergence. Due to its simplicity we baptize our model SGN.

The proposed SGN model is presented in the next Section. In Section 3 we present simulation results; in Section 4 we summarize our work and make a few concluding remarks.

## 2 SGN Proposed Model

Figure 1 describes our proposed architecture. Samples of a random vector $\mathbf{z} \in R^D$ are generated by a multivariate 'simple' (i.e. easily synthesized, typically Gaussian, Uniform, etc.) distribution $p_z(\mathbf{z})$. The generated samples are mapped by means of a parameterized function $F(\mathbf{z} ; \mathbf{w})$ into $\mathbf{y} \in R^d$, where D and d are the function F domain and range spaces dimensions respectively, and $\mathbf{w}$ is its parameters vector. The map $F: R^D \rightarrow R^d$ induces a corresponding implicit mapping between the distributions $p_z(\mathbf{z})$ and $p_y(\mathbf{y})$[1].

The system has also access to the data $\mathbf{x} \in R^d$ with same dimensionality as the generated samples $\mathbf{y}$. The parameterized function F naturally lends itself to a simple multi-layer feed forward neural network implementation. The network may be trained to solve the optimization problem:

$$w^* = \underset{w}{\operatorname{argmin}}\ d(p_y(x; w), p_x(x)) \qquad (1)$$

where d( . , .) is a suitable measure of discrepancy between both distributions, and where we made explicit, for clarity, the implicit effect of w upon $p_y(\mathbf{x})$. As it turns out the well known Kullback–Leibler divergence (a.k.a. KL-divergence)

---

[1] This mapping can in some cases be explicitly calculated and manipulated; in fact this precisely is what the fore-mentioned Normalizing Flows approach (Rezende and Mohamed, 2015) do.

$$D_{KL} = E_x\left[\log\left(\frac{p_x(x)}{p_y(x)}\right)\right] \quad (2)$$

may well serve such a measure. In conjunction with an appropriate density estimator it can be readily adapted to a stochastic gradient descent (SGD) scheme.

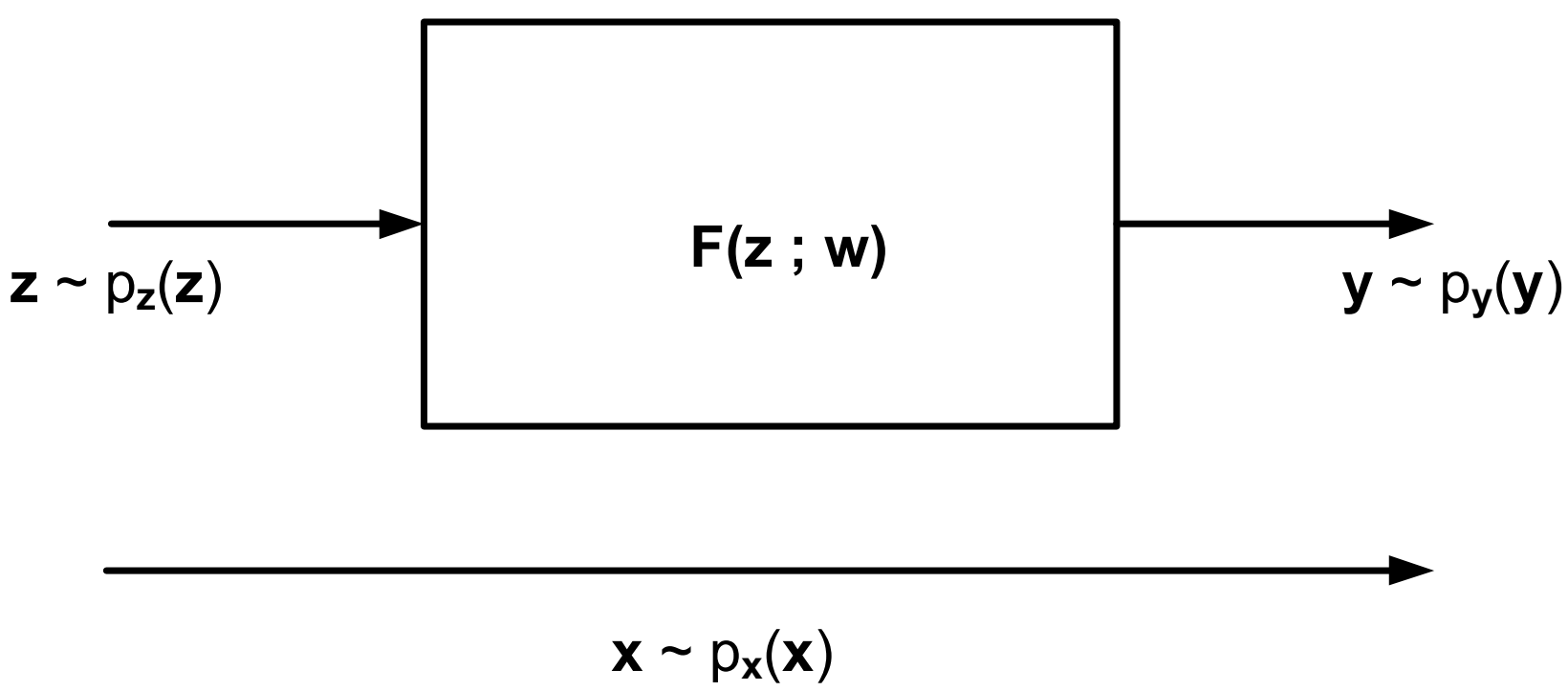

**Figure 1**: Proposed model architecture

More significantly, and in contrast with e.g. the popular log-likelihood optimization criterion, the KL-divergence is a bounded (from below) goal and thus provides a solid minimization convergence target. The expectation operator $E_{\mathbf{x}}[\ .\ ]$ (w.r.t. the data $\mathbf{x}$) can be approximated by application of the frequentist interpretation

$$D_{KL} \cong \frac{1}{M}\sum_{i=1}^{M}\log\frac{p_x(x_i)}{p_y(x_i)} \equiv \frac{1}{M}\sum_{i=1}^{M}D_{KL}^{i} \quad (3)$$

where $\{x_i\}$, i = 1, 2, …M is our data sample and $D_{KL}^{i}$ is *defined* as our SGD optimization goal.

Of course we do not have direct access to the densities $p_{\mathbf{x}}$ and $p_{\mathbf{y}}$ and we should estimate them. Of the available non-parametric estimators those based on the k-Nearest Neighbor (kNN ahead) method are especially suitable: they provide good quality density estimates of multivariate random vectors at both high and low density regions; they have a single not too sensitive parameter (k); and they are compactly differentiable. These kNN based estimators have been considerably studied (Wang, Kulkarni and Verdu, 2006; Wang, Kulkarni and Verdu, 2009).

A naïve implementation of the k-th nearest neighbor calculation (which is required for these estimators) conducts exhaustive (i.e. O(N)) search over a set such as $\{\mathbf{x}_l\}$, l = 1,2, …N of vectors and may become prohibitive for large samples such as those required by e.g. ImageNet (Deng et al., 2009). Many approximate kNN calculation methods exist, however; see e.g. (Li et al., 2016) for a comprehensive survey. Those of interest to us exhibit high performance, with an achievable average recall ratio of 0.8 (recall ratio is defined as the mean ratio between the number of found *true* nearest neighbors and k). And, more importantly, they incur O(log(N)) computational complexity for vectors of high dimensionality (e.g. Malkov and Yashunin, 2016). These methods potentially open the way to real life applications.

Following these considerations we pick (Wang, Kulkarni and Verdu, 2006) density estimator resulting in

$$\hat{p}_x(\mathbf{x}_i) = \frac{k}{(N\text{-}1)\, c(d)\, \rho_k^d(\mathbf{x}_i)} \tag{4a}$$

and

$$\hat{p}_y(\mathbf{x}_i) = \frac{k}{N\, c(d)\, \nu_k^d(\mathbf{x}_i)} \tag{4b}$$

where $\rho_k(\mathbf{x}_i)$ is the $L^2$ distance between $\mathbf{x}_i$ and its k-nearest neighbor in the data sample $\{\mathbf{x}_l\}\backslash\mathbf{x}_i$, denoted $\mathbf{x}_n$, $\nu_k(\mathbf{x}_i)$ is the $L^2$ distance between $\mathbf{x}_i$ and its k-nearest neighbor in the generated sample $\{\mathbf{y}_l\}$, denoted $\mathbf{y}_m$, $c(d) = \pi^{d/2} / \Gamma(d/2 + 1)$ is the d-dimensional unit ball volume, and $\Gamma(.)$ is the gamma function. The decrease of the densities in Equations (4a) and (4b) above with the increase of the distances $\rho_k(\mathbf{x}_i)$ and $\nu_k(\mathbf{x}_i)$ is of course intuitive. The difference between the 'N factors' in the equations reflects the fact that the search set $\{\mathbf{x}_l\}$ in (4a) should exclude $\mathbf{x}_i$ itself.

Applying Equations (4a) and (4b) into (3) we immediately get

$$D_{KL}^{i} = d \log\frac{\nu_k(\mathbf{x}_i)}{\rho_k(\mathbf{x}_i)} + \log\frac{N}{(N\text{-}1)} = \frac{d}{2}\log\frac{\|\mathbf{x}_i - \mathbf{y}_m\|^2}{\|\mathbf{x}_i - \mathbf{x}_n\|^2} + \log\frac{N}{(N\text{-}1)} \tag{5}$$

where $\| \; . \; \|$ denotes the $L^2$ norm; note that the dependence of $D_{KL}^{i}$ on k is now implicit through its impact on these norms. For our minimization purpose we may ignore the first factor and the last term in the right hand sides of (5), and provide an explicit expression for our simplified sample-wise SGD goal $D^i$

$$D^i = \log \frac{\sum_1^d (y_m^j - x_i^j)^2}{\sum_1^d (x_n^j - x_i^j)^2} \tag{6}$$

where j = 1,2,…d, and i = 1,2, …N.

We may notice by differentiating our goal $D^i$ that its derivative does not depend on $\mathbf{x}_n$, the k-nearest neighbor in $\{\mathbf{x}_l\}\backslash\mathbf{x}_i$ to the current sample $\mathbf{x}_i$ and so its calculation maybe skipped (in practice however we may wish to follow $D^i$ convergence plot; in such case $\mathbf{x}_n$ should obviously be required and calculated).

Following common practice we update the **w** parameter vector after processing a mini-batch comprised of B $\mathbf{x}_i$ samples. After each such **w** update a feed-forward operation of a full $\mathbf{y}_l = F(\mathbf{z}_l; \mathbf{w})$, l = 1,2, …N sample (using the updated **w** value) should be executed. This is required in order that further k nearest neighbor calculations be up-to-date. This creates a processing overhead. To maintain this overhead reasonably small the ratio N/B should be moderately small; since feed forward execution is much faster than back-propagation (e.g. ~ x400 faster on a Matlab based platform) a ratio N/B < 400 should apparently suffice.

Finally we note that contrary to 'conventional' neural nets (e.g. Classifiers) where differentiation of the goal function at sample $\mathbf{x}_i$ is executed w.r.t. the output at same corresponding sample (i.e. $\mathbf{y}_i$), in our case differentiation of the goal at data sample $\mathbf{x}_i$ is performed w.r.t. the network output $\mathbf{y}_m$ (that is the point in the set $\{\mathbf{y}_l\}$ k-nearest to $\mathbf{x}_i$).

# 3 Simulation Results

In order to probe and demonstrate the capabilities of this simple model a Matlab based platform was coded and tested over the MNIST dataset (LeCun et al., 1998). We built a small neural network with a single hidden layer of dimension 300. The generating random vector $\mathbf{z}$ was of dimension D = 100 and the output generated vector $\mathbf{y}$, compatible with the MNIST data $\mathbf{x}$, of dimension d = 28^2 = 784.

We experimented with both Uniform noise $\mathbf{z} \sim U[0, 1]^D$ and a Gaussian Mixture source (GM ahead) $\mathbf{z} \sim \sum_1^C p_i .N(\boldsymbol{\mu}_i, \sigma I)$ with C = 32, equal priors $p_i = 1/C$, means $\boldsymbol{\mu}_i$ drawn at setup from a binary random vector (of dimension D), i = 1,2,…C, $\sigma = 0.1$, and where I is the identity matrix.

The hidden layer activation function was a.tanh(.) with a = 5. Smaller values of a as well as use of ReLU activation (instead of tanh) resulted in generated samples with degraded contrast.

Random permutation of the MNIST training set was executed at the end of every training epoch (60000 characters).

We worked mainly with sample size N = 1e3 and mini-batch size B = 1e3 (the fore-mentioned ratio N/B was thus 1). We also performed experiments with N = 5e3 (and B = 5e3 too); since the larger the sample size N the better it reflects the actual data distribution this yielded as expected better MLL values (see ahead).

We used k = 1 as our kNN parameter; testing with k = 2 actually worsened our results. We used an exact exhaustive search scheme (of O(N) computational and memory complexity) for kNN; as mentioned above good quality approximate kNN methods of O(log(N)) computational complexity exist, such as (Malkov and Yashunin, 2016) and should be used with larger datasets maintaining an N/B ratio < 400; for example N = 1e5, B = 1e3, with N/B = 100 could possibly work out well on ImageNet.

We used SGD with no acceleration methods, with constant learning step $\eta = 0.2$ and with normalized goal function derivative.

Following the steps of many (Goodfellow et al., 2014; Li et al., 2015; Bengio et al., 2014; etc.) we use a protocol based upon a Gaussian Parzen Window density estimator in order to quantitatively evaluate our results and be able to compare them with other works. The estimator comprises 1e4 generated $\mathbf{y}$ points; the Gaussian Parzen

Window 'bandwidth' parameter was fine tuned by line search using a validation set of 5e3 points extracted from the MNIST training dataset. The empirical MLL of the 1e4 MNIST test dataset **x** points was calculated and reported. This method has been shown to yield unreliable results in some cases (Theis, van den Oord and Bethge, 2016; Wu et al., 2017) but no better quantitative method seems to have been generally adopted.

| MODEL | MLL | BY |
|---|---|---|
| Deep GSN | 214 | Bengio et al., 2014 |
| GAN | 225 | Goodfellow et al., 2014 |
| MMD (GMMN) | 147 | Li et al., 2015 |
| MMD (GMMN+AE) | 282 | " |
| | | |
| SGN ($\mathbf{z} \sim U[0,1]^D$, N = 5e3) | 244 | This work |
| SGN ($\mathbf{z} \sim U[0,1]^D$, N = 1e3) | 238 | " |
| SGN ($\mathbf{z}$ ~GM, N = 5e3) | 232 | " |
| SGN ($\mathbf{z}$ ~GM, N = 1e3) | 225 | " |

**Table 1**: Mean Log-Likelihood figures of merit

Please refer to Table 1 which describes MNIST MLL values of our work as compared with several generative models, all calculated using the above sketched methodology. SGN Uniform and GM noise sources parameters values are as described above. Simulation conditions at the cited works, such as source noise class and dimension, are either missing (e.g. GAN, Goodfellow et al., 2014) or incomplete (e.g. MMD, Li et al., 2015); hence we should treat the results in Table 1 as indicative rather than definitive. We notice that SGN yields competitive MLL values re cited state of the art methods. In particular SGN with Uniform source yields much better results (244) than GAN (225), and it is also strikingly superior to 'plain' (i.e. GMMN) MMD (147). This is, as mentioned above, the MMD scheme which shares with us a simple single neural net architecture. It might be of interest to implement an 'enhanced' SGN scheme (i.e. SGN + AE as the analog to the GMMN + AE variant (Li et al., 2015)) and check whether a significant quality boost is achieved, as in their case. We observe as expected, somewhat degraded MLL results as we decrease the sample size from N = 5e3 to N = 1e3 in both Uniform and GM sources.

We refer now to Figure 2, which shows a convergence plot of averaged $D^i$ (Equation (6) above) for a Gaussian Mixture source and sample size N = 1e3, with parameters values as described above. We can see that while initial convergence is rapid, final stages exhibit a 'long tail': it took approximately 2000 epochs to achieve the results shown in Table 1 and Figure 3b.

While true KL-divergence is bounded from below by 0 we notice convergence of our averaged $D^i$ to approximately -0.35. This is not to be attributed to our simplification from $D^i_{KL}$ (Equation (5) above) to $D^i$ (Equation (6)) but rather to the KL-divergence estimator bias under finite sample size (this estimator bias is known to only *asymptotically* vanish, see Wang, Kulkarni and Verdu, 2009). In fact we have experienced as expected a clear bias (absolute value) decrease (to about -0.16) upon increasing sample size to N = 1e4.

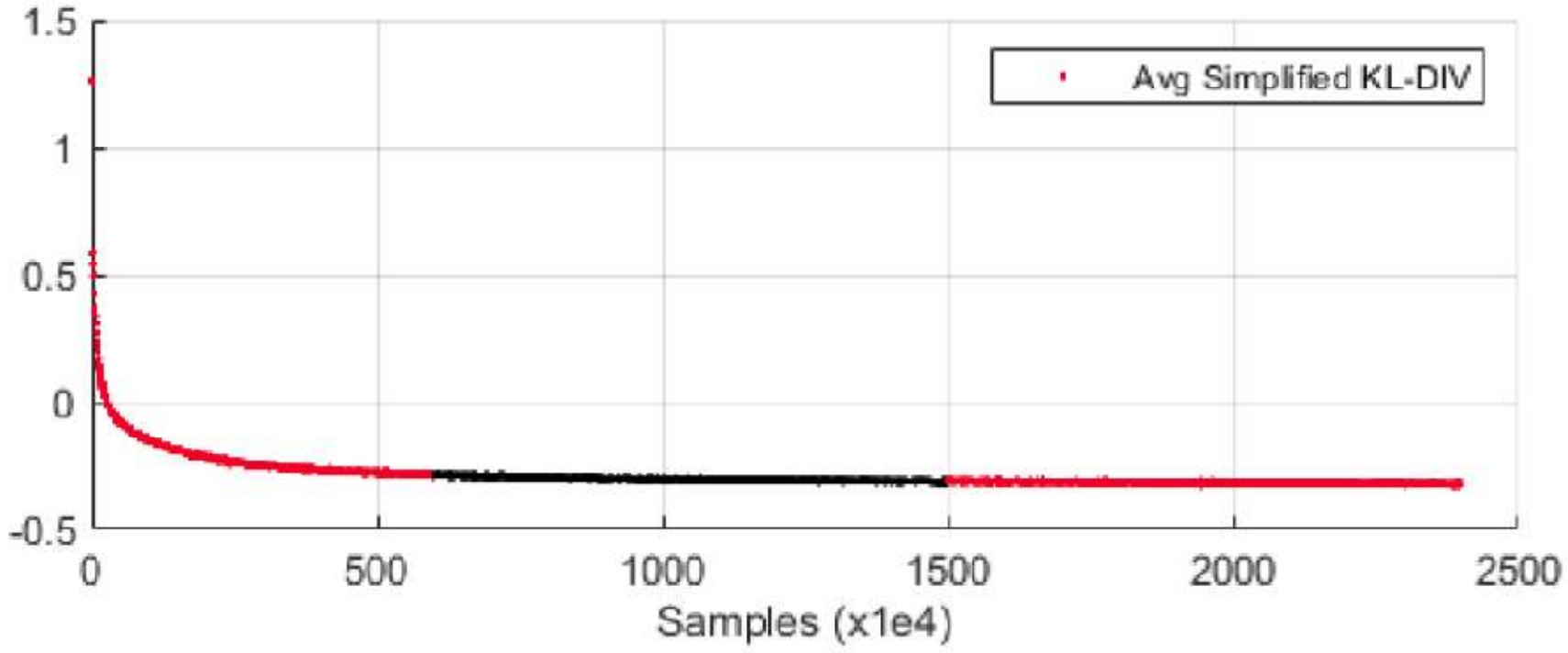

**Figure 2**: Averaged $D^i$ Convergence plot, GM source, N = 1e3; plot colors illustrate initial, interim and final convergence stages

Figure 3a provides visual samples of the MNIST data **x** for reference. Figure 3b shows a random set of samples generated by a GM distribution source and sample size N = 1e3 (all other parameters as described above). It may be verified that all digits classes are represented. We leave for the reader to judge their visual quality. Similar visual quality is achieved by $U[0,1]^D$ with N = 1e3. Samples generated by GM and $U[0,1]^D$ sources with N = 5e3, appear to exhibit, again as expected, somewhat better visual quality.

Figure 3c shows the result of interpolating, in latent space, from a randomly selected corner point of the $U[0,1]^D$ hypercube to the opposite corner along a major diagonal; near one corner the digit '7' predominates, smoothly passing through '8' at the center region of the hypercube and ending at '6' at the opposite corner.

It is interesting to generate samples from a Gaussian or Uniform Mixture source, with the number of modes at least equal to the number of distinct classes and with sufficiently separated modes (i.e. small mode dispersion relative to the intermodal distances). One might intuitively expect in such a case that each mode will generate samples belonging to one and only one class. Interpolation results such as those shown in Figure 3c seem to intuitively support this idea. Should this be the case it would be perhaps possible to achieve good (that is linearly separable) unsupervisedly learned representations, by means of e.g. complementing SGN to an auto-encoding scheme. This is work in progress, and no results to report yet.

## 4 Concluding Remarks

A generative network model, denoted SGN, consisting of a multi-layer feed-forward neural net along with a kNN based KL-divergence goal function was presented. Given its (practically trivial) simplicity it is surprising that such a model appears to have been overlooked. When tested on MNIST data it was shown to generate visually appealing samples with competitive mean log-likelihood metrics relative to state of the art methods.
To extend its applicability to more realistic datasets (such as ImageNet) a more computationally efficient kNN method should be implemented than the one, of complexity O(N), used herein. A significant variety of such methods, suitable to high dimensionality spaces, have been studied with approximate but impressive performance. Some of them exhibit O(log(N)) complexity. The integration of such a method within the SGN concept, as well as the assessment of the impact of its approximate nature upon the quality of the generated samples (relative to the exact kNN method) is left for future work.

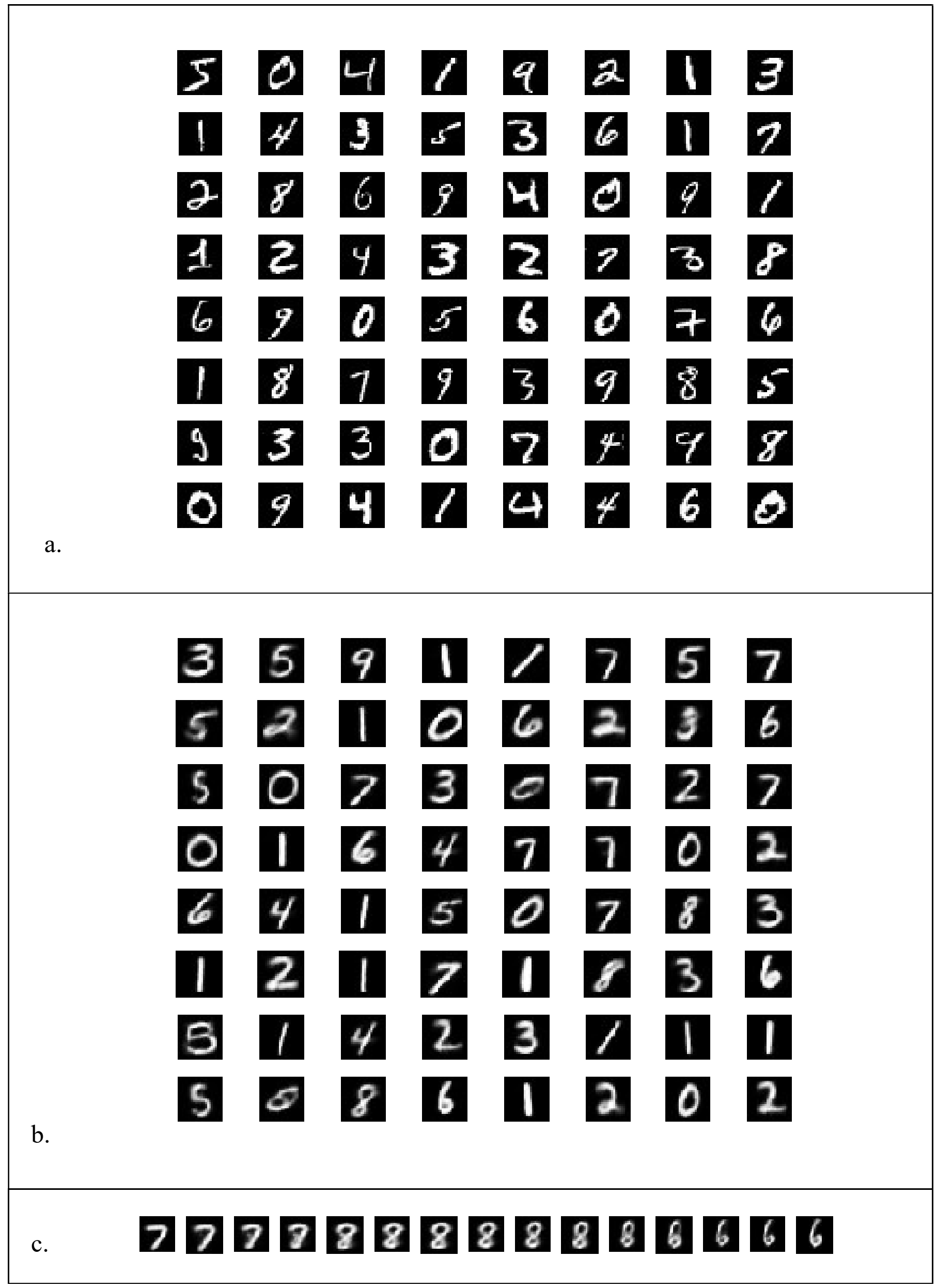

**Figure 3**: a. MNIST samples; b. GM, N = 1e3 generated samples; c. $U[0,1]^D$, N = 1e3 Latent space interpolated samples

Due to the poor results of the MMD (GMMN) method with a simple architecture (like ours), the authors turned their focus to a more complex, auto-encoding enhanced scheme, and achieved an impressive performance boost, both in terms of visual quality and MLL. It would be interesting to test whether such combined architecture has a similar boosting effect on SGN.

A Matlab based SGN software package will be provided upon request.

## References

[1] Bengio, Y., Thibodeau-Laufer, E., Alain, G., 'Deep Generative Stochastic Networks Trainable by Backprop', ICML 2014

[2] Bond-Taylor, S., Leach, A., Long, Y., Willcoks, C.G., 'Deep Generative Modeling: A Comparative Review of VAEs, GANs, Normalized Flows, Energy-Based and Autoregressive Models', arXiv 2103.04922, 2021

[3] Deng, J. Dong, W., Socher, R., Li, L., Li, K., Fei-Fei, L., 'ImageNet: A Large-Scale Hierarchical Image Database', CVPR 2009

[4] Dziugaite, K.D., Roy, D.M., Ghahramani, Z., 'Training generative neural networks via Maximum Mean Discrepancy optimization', UAI 2015

[5] Goodfellow, I., Pouget-Abadie, J., Mirza, M., Xu, B., Warde-Farley, D., Ozair, S., Courville, A., Bengio, 'Generative Adversarial Networks', NIPS 2014

[6] Kingma, D.P., Welling, M., 'Auto-Encoding Variational Bayes', ICLR 2014

[7] LeCun, Y., Bottou, L., Bengio, Y., Haffner, P., 'Gradient-Based Learning Applied to Document Recognition', Proc. of the IEEE, 1998

[8] Li, Y. Swersky, K., Zemel, R., 'Generative Moment Matching Networks', ICML 2015

[9] Li, W., Zhang, Y., Sun, Y., Wang, W., Zhang, W., Lin, X., 'Approximate Nearest Neighbor Search on High Dimensional Data --- Experiments, Analyses, and Improvement', IEEE Trans. On Knowledge and Data Engineering, 2020

[10] Malkov, Y.A., ' and Yashunin, D.A., 'Efficient and Robust Approximate Nearest Neighborr Search Using Hierarchical Navigable Small World Graphs', IEEE Trans. On PAMI, 2016

[11] Mnatsakanov, R.M., 'Hausdorff moment problem: Reconstruction of probability density functions', Statistics and Probability Letters, 2008

[12] Mohamed, S. and Lakshminarayanan, B., 'Learning in Implicit Generative Models', ICLR 2017

[13] Rezende, D.J., Mohamed, S., 'Variational Inference with Normalizing Flows', ICML 2015

[14] Theis, L., van den Oord, A., Bethge, M., 'A Note on the Evaluation of Generative Models', ICLR 2016

[15] Wang, Q., Kulkarni, S.R., Verdu, S., 'A Nearest-Neighbor Approach to Estimating Divergence between Continuous Random Vectors', ISIT 2006

[16] Wang, Q., Kulkarni, S.R., Verdu, S., 'Divergence Estimation for Multidiimensional Densities via k-Nearest-Neighbor Distances', IEEE Trans. On IT, 2009

[17] Wu, Y., Burda, Y., Salakhutdinov, R., Grosse, R., 'On the Quantitative Analysis of Decoder-based Generative Models', ICLR 2017